\documentclass{article}

\usepackage[final]{corl_2017} 
\usepackage{amsmath}
\usepackage{graphicx,wrapfig,lipsum}
\usepackage{subcaption}
\usepackage{optidef}
\usepackage{drum}

\title{Learning Data-Efficient Rigid-Body Contact Models: Case Study of Planar Impact}

\author{
    Nima Fazeli\\
    Department of Mechanical Engineering\\
    Massachusetts Institute of Technology
    MA, United States\\
    \texttt{nfazeli@mit.edu} \\
    \And
    Samuel Zapolsky \\
    Toyota Research Institute \\
    CA, United States \\
    \texttt{sam.zapolsky@tri.global} \\
    \AND
    Evan Drumwright \\
    Toyota Research Institute \\
    CA, United States \\
    \texttt{evan.drumwright@tri.global} \\
    \And
    Alberto Rodriguez \\
    Department of Mechanical Engineering\\
    Massachusetts Institute of Technology
    MA, United States\\
    \texttt{albertor@mit.edu} \\
}

\begin{document}
\maketitle


\begin{abstract}
In this paper we demonstrate the limitations of common rigid-body contact models used in the robotics community by comparing them to a collection of data-driven and data-reinforced models that exploit underlying structure inspired by the rigid contact  paradigm. We evaluate and compare the analytical and data-driven contact models on an empirical planar impact data-set, and show that the learned models are able to outperform their analytical counterparts with a small training set.
\end{abstract}

\keywords{Contact Modeling, Rigid-Body Dynamics, Data-Efficient Learning} 

\section{Introduction} \label{sec:intro}
Rigid-body contact models are ubiquitous in robotics with applications, for example, in planning (\citet{Posa2013}), control (\citet{Hogan2016}), system identification (\citet{fazeli2016parameter,Fazeli2016,Zhou2016}), and state-estimation (\citet{Koval2013,Koval2015}). In this paper we briefly review six established rigid contact models from the literature, and evaluate the predictive performance of these models on an empirical data-set. Next we highlight the limitations of these models as motivation for the need to involve data in the modeling process; and propose two classes of data-efficient learning paradigms inspired by rigid contact models and show that they are able to outperform their counterparts.

Prevalent contact models used in robotics assume rigid body dynamics and can generally be categorized as: i) pseudo-rigid models (i.e., a thin and inertialess compliant layer over a rigid core)  and ii) rigid contact models (resulting from constraint equations). These models have the desired properties of computational efficiency and simplicity, as opposed to finite element models that are expensive to compute and not suitable for realtime applications. Penalty methods assume that the bodies in contact do not deform but allow for penetration as a proxy to deformation, and compute the imparted contact forces as a function of the amount of penetration. Adams multi-body simulator \cite{li2001virtual} (commercialized) and Dance \cite{NgThowHing1999DANCEDA} (open source) are penalty method simulation packages. Akin to these penalty methods are the relaxations via regularization approaches, as deployed in \citet{todorov2012mujoco}. 

Rigid-body dynamic simulation using the complementarity formulation as popularized by \citet{Anitescu:1997} and \citet{Stewart:1996} resolves contact without penetration, assumes perfectly rigid bodies, and treats contact as a instantaneous and discontinuous event. These approaches make use of contact models (such as \citet{Stronge:1990,Wang:1992,Whittaker:1944}) that are amenable to the complementarity formulation. Regarding both categories, only a handful of studies (\citet{Fazeli2017contact,Fazeli2017isrr}) have tried to provide formal empirical verification of contact models and model selection is treated mostly as a matter of convenience, highlighting the need for experimental work and data-driven approaches.

Motivated by the modicum of empirical work in contact model verification and selection, as well as the limited usage of data in improving and constructing effective contact models (\citet{Yu2016}), we first review several common rigid body contact models and show their limitations in making predictions, then construct data-driven models using the underlying rigid-body contact assumption and demonstrate that they are able to out-perform the best possible predictions made by the models.


\section{Background - Rigid-Body Contact Models} \label{background}

\begin{figure}
    \centering
    \includegraphics[width=0.95\textwidth]{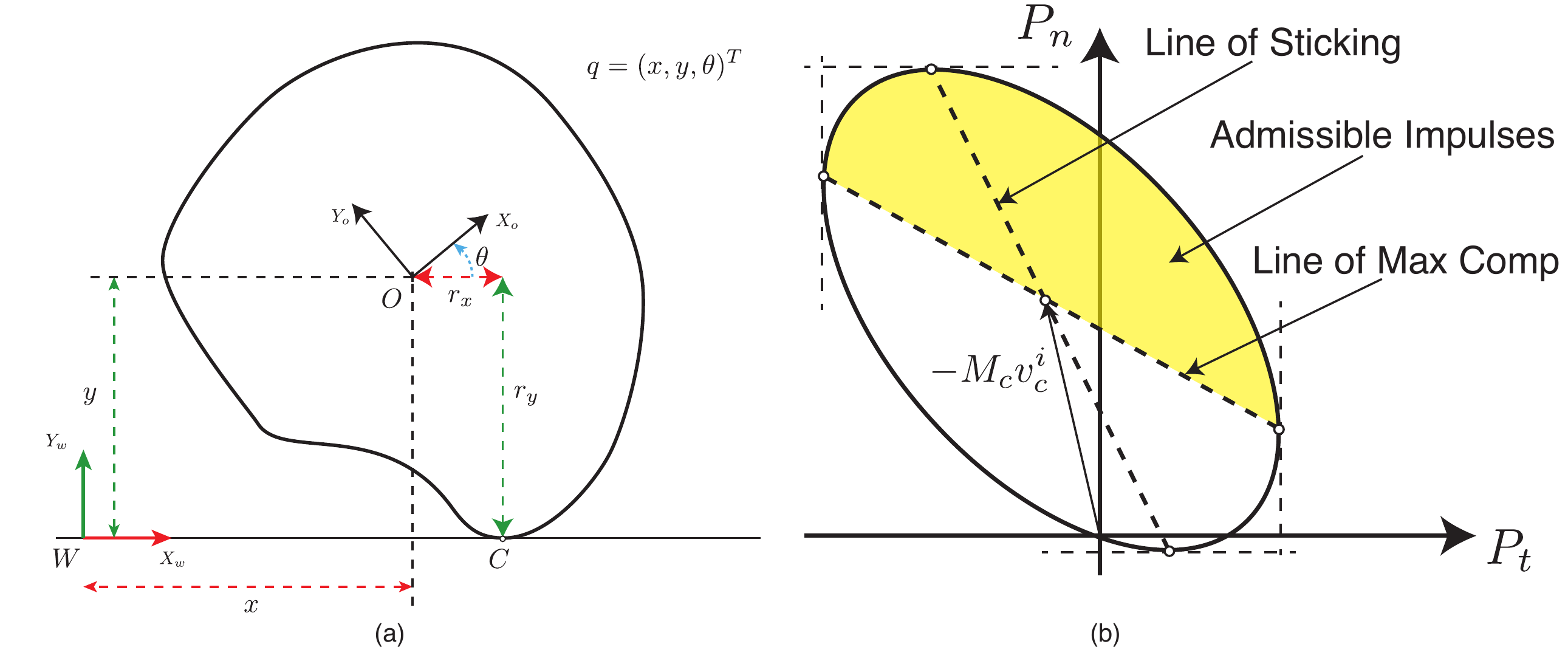}
    \caption{a) planar object making contact with a fixed horizontal surface, b) Energy Ellipse.}
    \label{fig:2.1}
\end{figure}

Consider Fig.~\ref{fig:2.1}a, where a planar object makes contact with a horizontal surface, the purpose of a contact model is to predict the imparted impulses and the resultant states of the object. We use $m$ and $I$ to represent the mass and second moment of inertia of the object, and $(r_x,r_y)$ denote the position of the contact point with respect to the center of mass of the object in an inertial frame. We represent the configuration of the object with $\vect{q}=(x,y,\theta)^T$, the velocity with $\vect{v}=\dot{\vect{q}}$, the imparted impulse during impact with $\vect{P}$, and use superscripts $i$ and $f$ to denote pre- and post-contact variables. A contact model performs the mapping:
\begin{align} \label{eq:2.1}
    \vect{f}_{contact}: \{m,I,r_x,r_y,\vect{q},\vect{v} \}^i_{10\times 1} \xrightarrow{\vect{P}} \{\vect{q},\vect{v}\}^f_{6\times 1}
\end{align}
This mapping does not have a functional form yet. Rigid-body contact models assume that deformations are negligible during impact, and resolve contact instantaneously. This assumption means that the configuration of the object pre- and post-contact does not change and the impact happens at a single point on the body. Point contacts permit only linear impulses to be transmitted between the bodies. We can write the dynamic equations of motion for the instant before and after the impact for a rigid-body as:
\begin{align} \label{eq:2.2}
    \vect{v}^f = \vect{v}^i + \mat{M}^{-1} \mat{J}^T \vect{P}
\end{align}
where $\mat{M}_{3\times 3}$ denotes the inertia matrix, $\mat{J}_{2\times 3}$ is the contact Jacobian, and $\vect{P}_{2\times 1}$ is the linear imparted impulse. The rigid-body assumption gives Eq.~\ref{eq:2.1} a functional form in Eq.~\ref{eq:2.2} but is still unresolvable since $\vect{P}$ and $\vect{v}^f$ are both unknowns.

The impulse $\vect{P}$ is constrained by the law of conservation of energy and that no penetration can occur. To derive a mathematical representation for these two constraints, we pre-multiply Eq.~\ref{eq:2.2} by $\mat{J}$:
\begin{align} \label{eq:2.3}
    \vect{v}_c^f = \vect{v}_c^i+ \mat{J}\mat{M}^{-1}\mat{J}^T \vect{P} \rightarrow \vect{v}_c^f = \vect{v}_c^i+ \mat{M}_c^{-1} \vect{P}
\end{align}
where $\vect{v}_c$ denotes the velocity of the contact point, and $\mat{M}_c$ represents the effective inertia at the point of contact. We may now express the conservation of kinetic energy for the contact point as:
\begin{align} \label{eq:2.4}
    (\vect{v}_c^f)^T \mat{M}_c \vect{v}_c = \alpha (\vect{v}_c^i)^T \mat{M}_c \vect{v}_c^i, \quad \alpha \in [0,1]
\end{align}
Plugging in Eq.~\ref{eq:2.3} we arrive at:
\begin{align} \label{eq:2.5}
    (\vect{P}+\mat{M}_c \vect{v}_c^i)^T \mat{M}_c^{-1} (\vect{P}+\mat{M}_c \vect{v}_c^i) = \alpha (\vect{v}_c^i)^T \mat{M}_c \vect{v}_c^i
\end{align}
Eq.~\ref{eq:2.5} defines an ellipse in the impulse space ($\vect{P}$), with an offset of $-\mat{M}_c \vect{v}_c^i$ which is the incident momentum of the contact point. This ellipse is called the Energy Ellipse and is explained in more detail in (\citet{chatterjee1997rigid,Fazeli2017isrr}). Fig.~\ref{fig:2.1}b shows an example of an Energy Ellipse, the line of maximum compression denotes impulses that yield zero separating velocities, and the line of sticking denotes the impulses that lead to a zero relative tangential velocity component. The lightly shaded region is the space of admissible impulses subject to the energy and non-pentration constraints. These constraints are still not sufficient to resolve the contact, contact models introduce additional constraints and parameters to select a unique imparted impulse from this admissible set.

\subsection{Analytical Contact Models Studied}
Analytical contact models proposed in the literature parameterize some region within the admissible impulse space, and select a unique impulse given pre-contact states and parameters. The selected impulse $\vect{P}$ is then plugged into the equations of motion (Eq.~\ref{eq:2.2}) to predict the velocity vector for the next time instant. 

In this study, we consider 6 commonly used analytical models: \textbf{1) Anitescu-Potra Newton} \cite{Anitescu:1997}, \textbf{2) Anitescu-Potra Poisson} \cite{Anitescu:1997}, \textbf{3) Drumwright-Shell} \cite{Drumwright:2010b}, \textbf{4) Mirtich} \cite{Mirtich:1996vt}, \textbf{5) Wang-Mason} \cite{Wang:1992}, and \textbf{6) Whittaker} \cite{Whittaker:1944}. The details of these models are summarized in \citet{Fazeli2017contact}. In addition to these models, we define two post hoc models ``\textbf{Best Post Hoc}" and ``\textbf{Ideal Rigid Body Bound (IRB Bound)}". The Best Post Hoc model chooses the analytical model that predicted the outcome most accurately for each trial (the best possible performance for the 6 models), and the IRB Bound chooses an impulse in the Energy Ellipse that best explains the trial given the outcome (establishing an absolute upper bound on the performance of any rigid contact model satisfying the standard assumptions made). We present the computation of the IRB Bound in sec.~\ref{sec:3.2}.







\begin{figure}[h]
    \centering
    \includegraphics[width=0.95\textwidth]{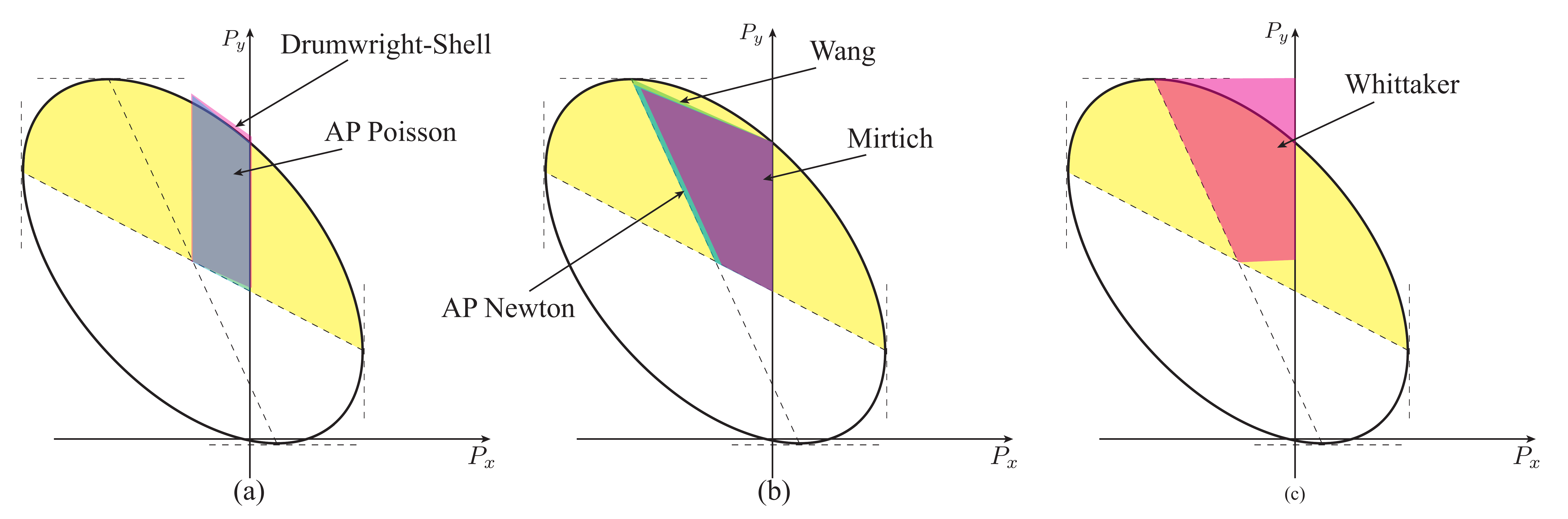}
    \caption{Space of admissible impulses demarcated by the contact models for all possible choices of parameters.}
    \label{fig:2.3}
\end{figure}

The key common feature of the six analytical models is that they are all characterized by two parameters, a normal restitution coefficient ($\epsilon$) and a tangential friction coefficient ($\mu$), though each model uses and interprets these parameters differently. The normal restitution regulates the magnitude of the imparted normal impulse (a larger value indicates a larger impulse and a larger ``bounce") and the tangential fiction coefficient regulates the tangential impulse, and discriminates between sticking and sliding during the impact event. Fig.~\ref{fig:2.3} shows the regions of predictions of the models within the Energy Ellipse for all possible values of the parameters, grouped based on graphical similarity. Though these regions appear similar graphically, they are constructed differently computationally. 

Fig.~\ref{fig:2.3} shows that no model is able to cover a large portion of the admissible space, and that the regions of predictions are quite similar. Further, we note that none of the six studied models is able to predict points to the opposite side of the line of sticking with respect to their own subspace. This phenomenon is referred to as ``back-spin" \cite{Chatterjee:1998} and occurs during our experiments. We can also infer the definitions of the parameters used in the models from the shape of the admissible regions, for example the Poisson restitution is a ratio of the normal momenta at contact, therefore the models that use it have vertical lateral boundaries. 

\section{Experiments and the Performance of the Analytical Models} \label{sec:experiments}
\subsection{Experimental Setup:}
\begin{figure}
    \centering
    \begin{subfigure}[b]{0.4\textwidth}
        \includegraphics[width=\textwidth]{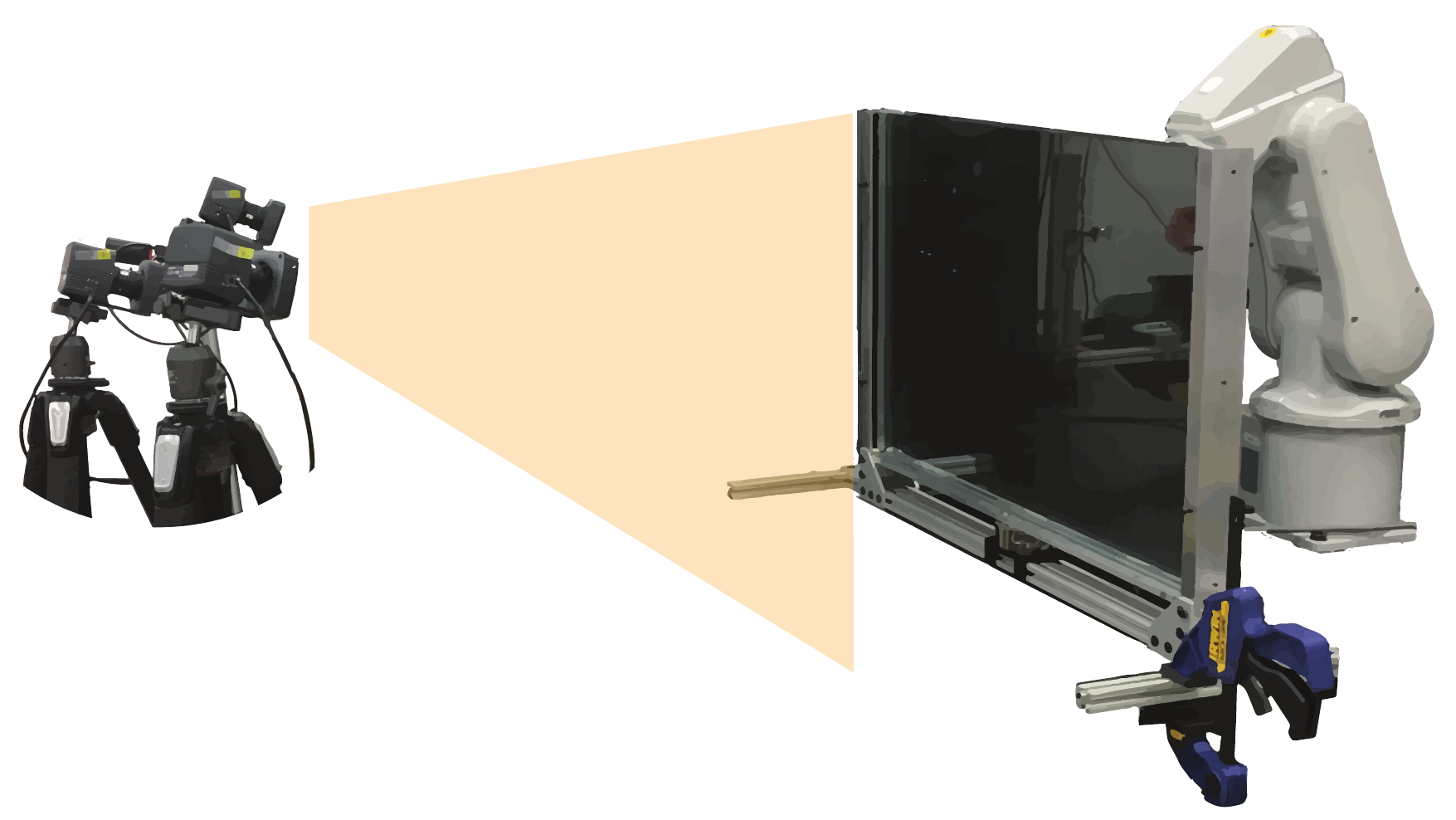}
        \caption{Planar impact experimental setup. The cameras record the motion of the object at 250 hz, and the robot imparts randomized initial conditions to the motion of the object in every trial.}
        \label{fig:3.1}
    \end{subfigure}
    ~ ~ ~ ~ ~ ~ ~ 
    \begin{subfigure}[b]{0.33\textwidth}
        \includegraphics[width=\textwidth]{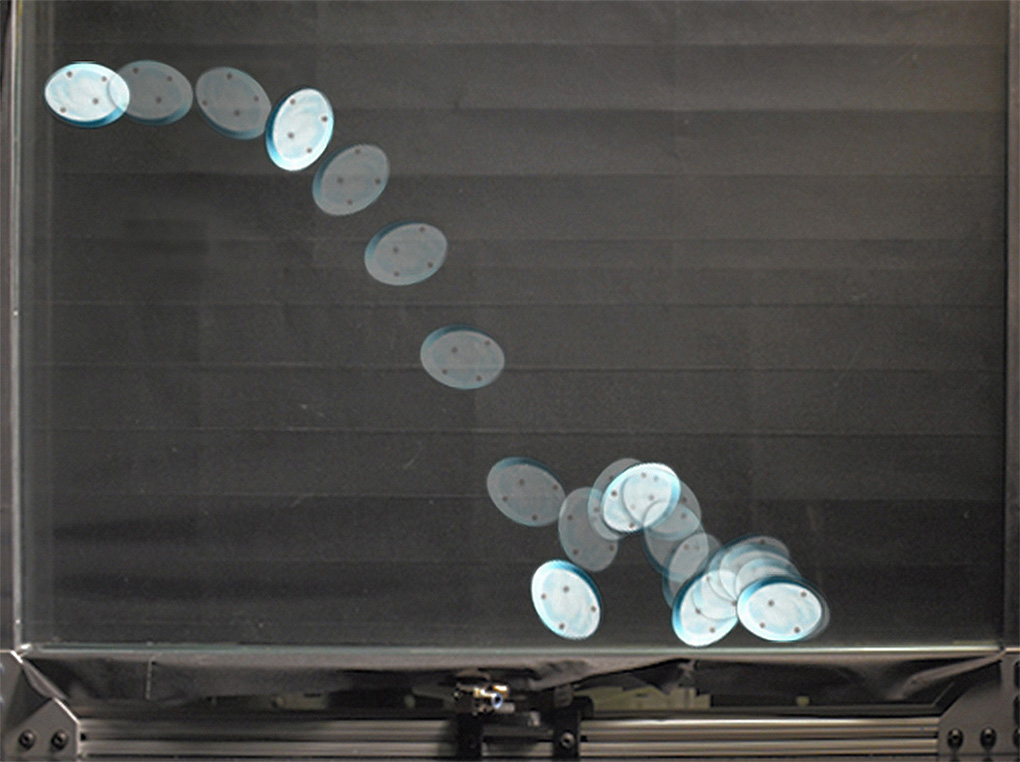}
        \caption{Planar object making contact with a fixed horizontal surface.}
        \label{fig:3.2}
    \end{subfigure}
    \caption{Experimental setup: robotic autonomous impact data collection.}
\end{figure}

We devised the experimental setup in Fig.~\ref{fig:3.1} for the evaluation and learning of the contact models. The experiment consists of planar impact of objects subject to gravity, and is designed for autonomous data collection with the robot. Fig.~\ref{fig:3.2} depicts an example trajectory of an ellipse.

The setup consists of a dropping arena, where two flat sheets of glass constrain the motion of planar objects to the vertical plane, the motion tracking cameras to measure the states of the object, and the robot with a directional magnetic latch mechanism to manipulate the object (equipped with similar directional magnets) between the glass sheets. In each run of the experiment, the robot generates a random set of initial conditions for the drop, the object is released with these conditions, and the subsequent motion the object is recorded. In total 2000 drops were generated and 1718 of them were selected after filtering for data integrity (missing frames, reflection, unexpected releases etc.). Next, contact events where detected, pre- and post-contact states where extracted, and the physical properties of the object where carefully measured. For more details please see \citet{Fazeli2017isrr}.

\subsection{Rigid-body Contact Model Predictive Performances}\label{sec:3.2}

\begin{figure}
    \centering
    \begin{subfigure}[b]{0.49\textwidth}
        \includegraphics[width=0.95\textwidth]{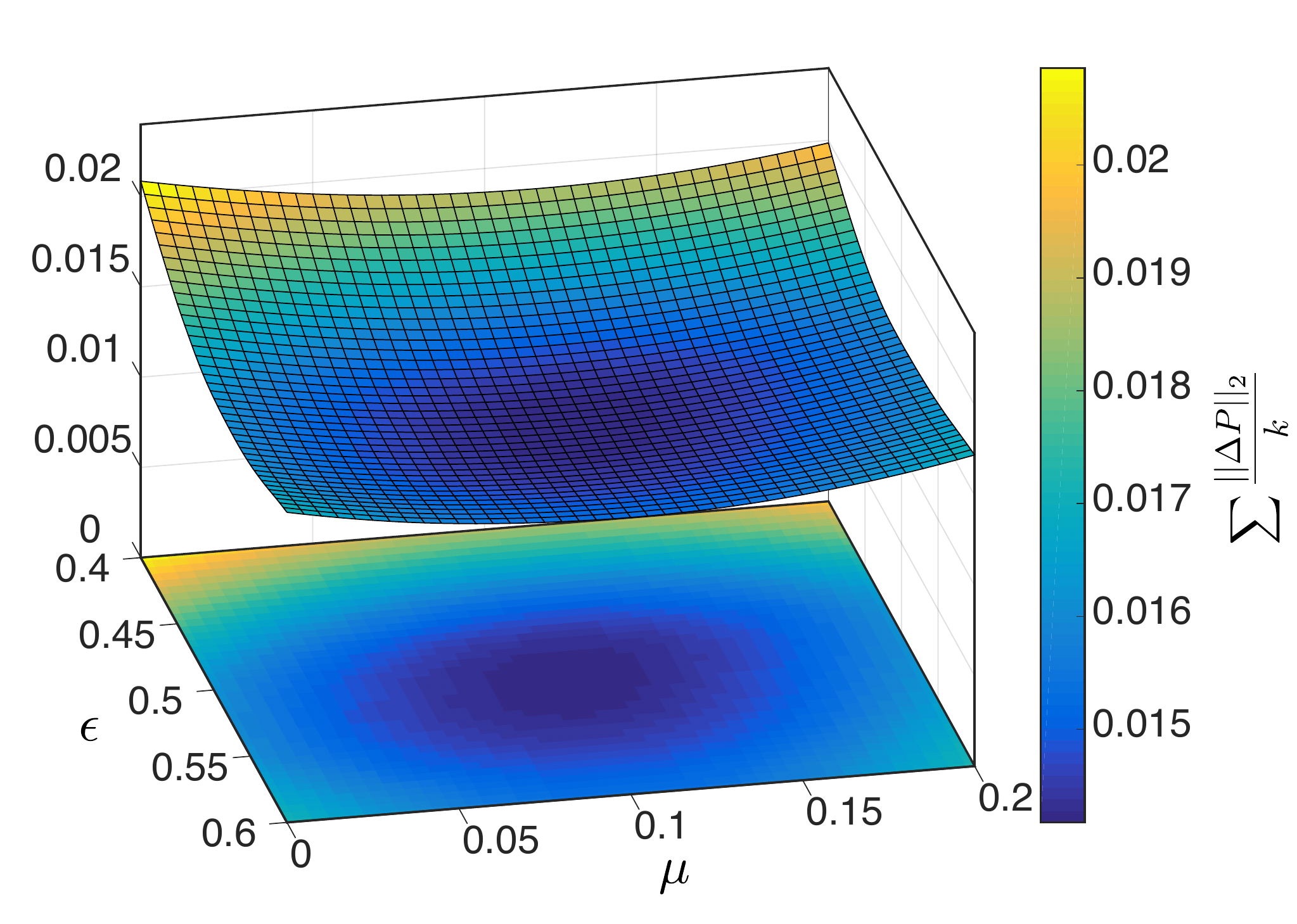}
        \caption{}
        \label{fig:3.3}
    \end{subfigure}
    ~
    \begin{subfigure}[b]{0.49\textwidth}
        \includegraphics[width=0.95\textwidth]{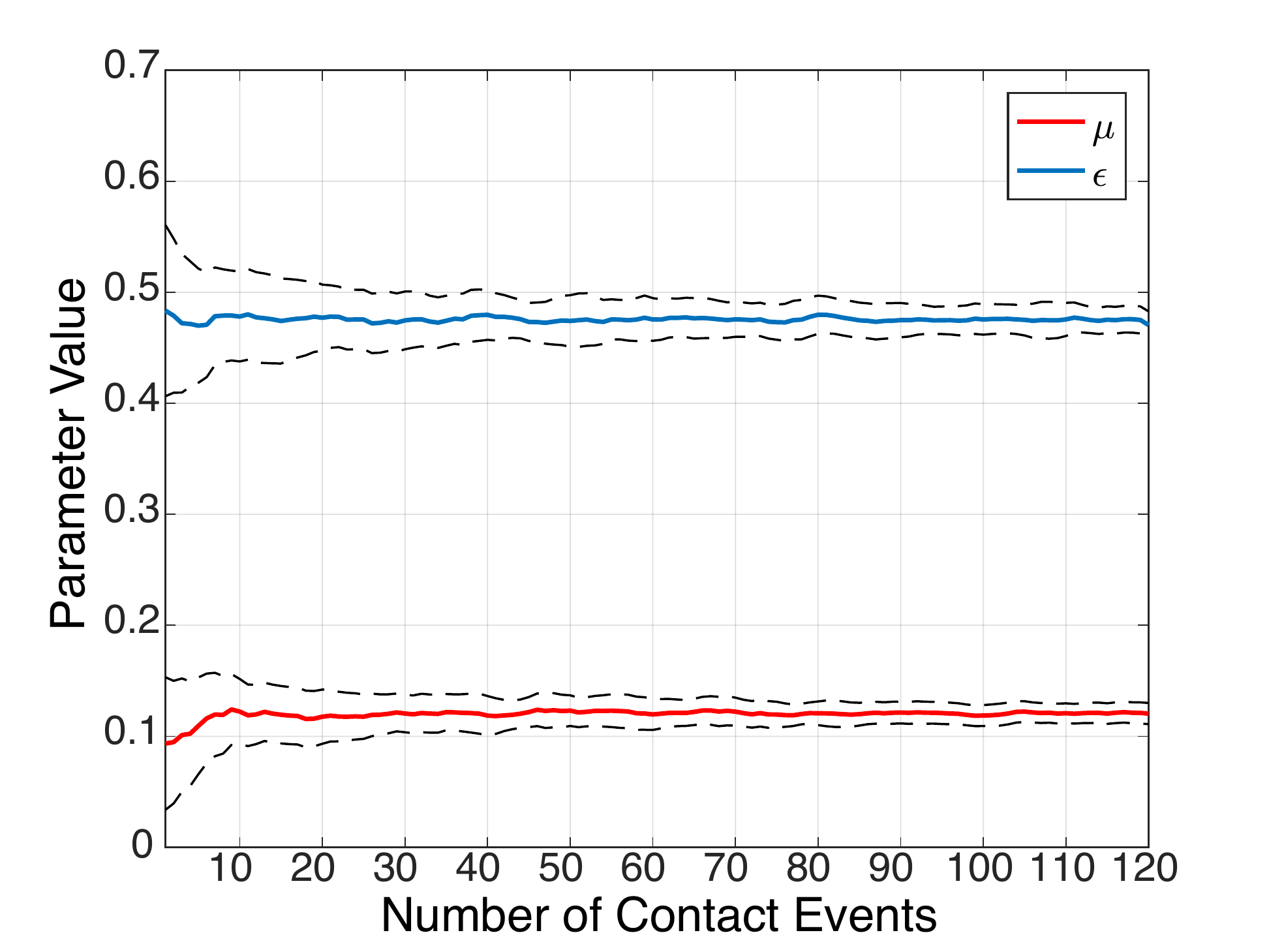}
        \caption{}
        \label{fig:3.4}
    \end{subfigure}
    \caption{a) Whittaker model objective function for $k=120$. b) Convergence of Whittaker model parameters as a function of number of trials. Solid colors show mean parameter value, and dashed black show 1 standard deviation.}
\end{figure}

An effective way of quantifying the predictive performance of the contact models is to compare the measured post contact velocities of the object with the values predicted by the models. To this end, first we find the optimal model parameters (system identification) for each model. For each model, we chose $k$ trials of the experiment without replacement $m$ times, and for each of $m$ iterations we find the optimal parameters by solving:
\begin{argmini} 
    {\mu,\epsilon}{\sum_{n=1}^k || \vect{P}_n - \hat{\vect{P}}_n ||_2}
    {\label{eq:3.1}}{}
    \addConstraint{\hat{\vect{P}}_n}{=\vect{f}_c(\mu,\epsilon,\vect{q}_n^i,\vect{v}_n^i)}
\end{argmini} 
Here $\vect{f}_c$ denotes the contact model of interest, $\vect{P}_n$ denotes the measured impulse for trial $n$, and $\hat{\vect{P}}_n$ is the estimated impulse from the contact model. Next we compute the average and standard deviation over the $m$ iterations. Fig.~\ref{fig:3.3} shows the surface of the cost function (Eq. \ref{eq:3.1}) plotted as a function of the parameters for the Whittaker model for $k=120$. Fig.~\ref{fig:3.4} shows the convergence behaviour of the model parameters for $k=1, \ldots, 120$. For more details on the identification procedure see \citet{Fazeli2017isrr}. Tab.~\ref{tab:3.1} shows the computed optimal model parameter values for each model.

\begin{table}[]
    \centering
    \caption{Optimal parameters for the models.}
    \setlength\tabcolsep{0.5cm}
    \begin{tabular}{| c || c | c ||} \hline 
        \textbf{Model}  &   $\mu$    & $\epsilon$ \\ \hline \hline
        DrumShell   &   $0.081 \pm 0.008$    & $0.516 \pm 0.009$      \\ 
        AP Poisson  &   $0.101 \pm 0.007$    & $0.526 \pm 0.008$       \\ 
        AP Newton   &   $0.084 \pm 0.007$    & $0.547 \pm 0.009$      \\ 
        Mirtich     &   $0.062 \pm 0.009$    & $0.558 \pm 0.010$       \\
        Wang-Mason  &   $0.120 \pm 0.008$    & $0.537 \pm 0.010$       \\
        Whittaker   &   $0.111 \pm 0.008$    & $0.484 \pm 0.011$      \\ \hline
    \end{tabular}
    \label{tab:3.1}
\end{table}

\begin{figure}
    \centering
    \includegraphics[width=0.5\textwidth]{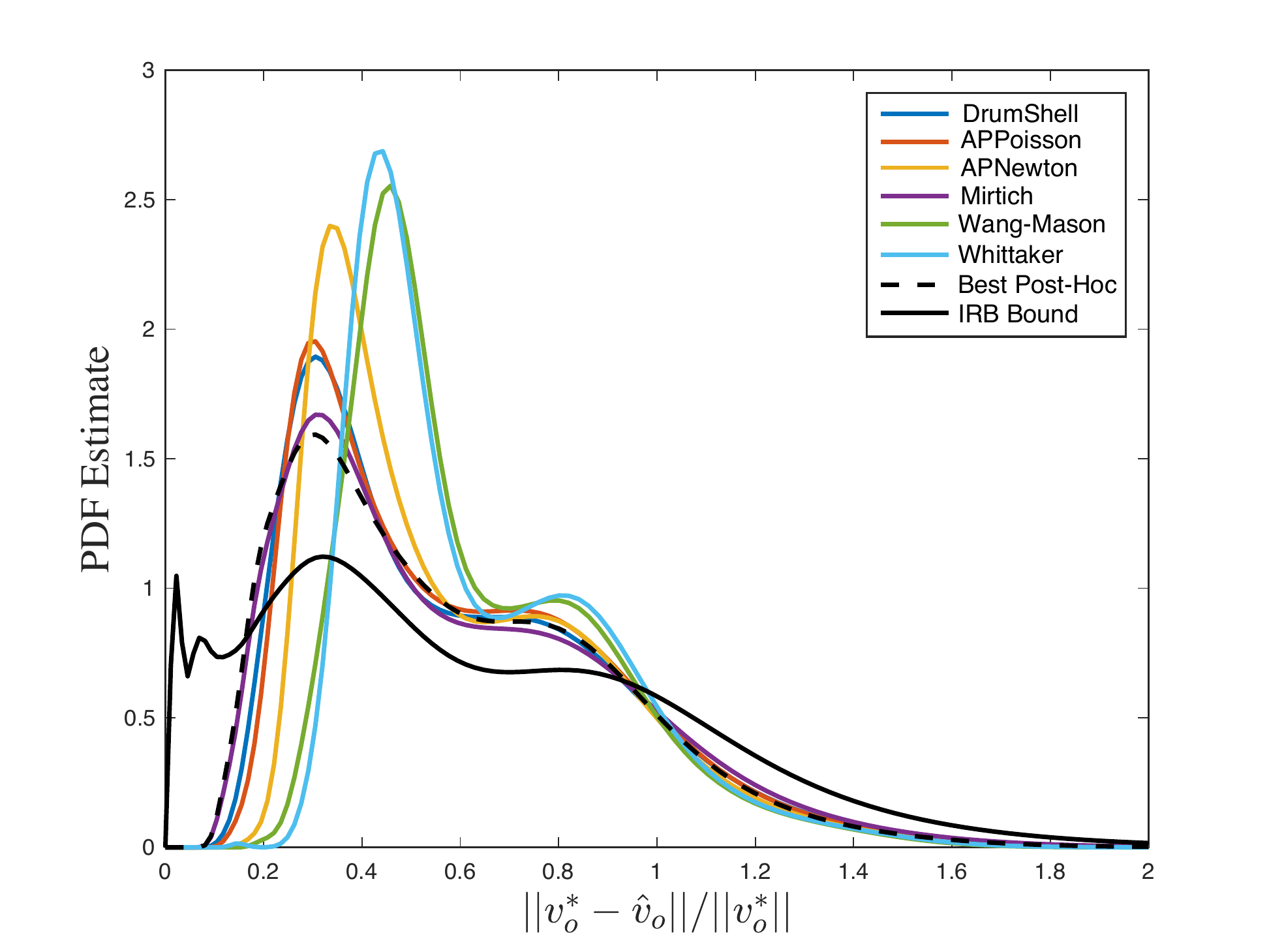}
    \caption{Normalized $l_2$ norm error in predicted post contact center of mass velocity.}
    \label{fig:3.5}
\end{figure}

With the identified optimal parameters we are ready to evaluate the predictive performance of the models, here we chose the $l_2$ norm of the error in predicted post-contact center of mass velocity of the object as the metric. Fig.~\ref{fig:3.5} shows the performance of the six analytical models and the two post hoc models. The IRB Bound is the post hoc model that selects the impulse in the admissible space of the Energy Ellipse that best explains the trial and is computed by solving:
\begin{argmini}
    {\vect{P}}{|| \vect{v}_n^f - \vect{v}_n^i - \mat{M}^{-1}\mat{J}^T \vect{P}||_2}
    {}{}
\end{argmini}
where $\vect{v}_{3\times 1}$ denotes the velocity of the center of mass of the object, and $\vect{P}_{2\times 1}$ is the estimated imparted impulse under the rigid-body contact model. All the contact models studied so far assume point contact and only allow a linear impulse to be imparted on the object and the IRB Bound gives an absolute bound on the performance of any model constructed from the rigid body contact model assumptions. In sec.~\ref{sec:data-driven-models} we learn models that relax some of these assumptions.

\begin{figure}
    \centering
    \includegraphics[width=\textwidth]{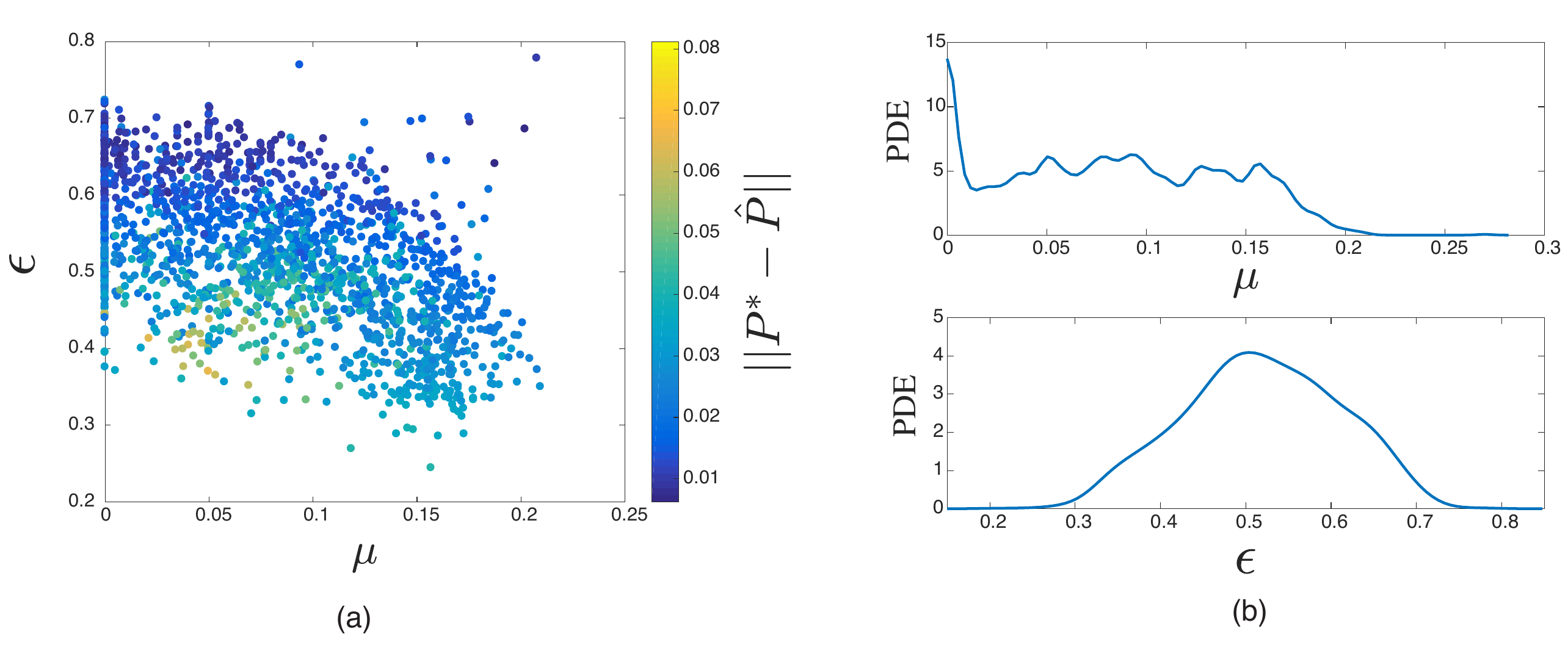}
    \caption{a) Anitescu-Potra Newton model parameters and corresponding error in predicted impulse, blue to yellow indicates increase in error, b) AP Newton model parameters $(\mu,\epsilon)$ distributions.}
    \label{fig:3.6}
\end{figure}

We also study each impact individually and identify the contact parameters for each impact that most optimally explain the event. Fig.~\ref{fig:3.6}a shows the $(\mu_k,\epsilon_k)$ for each drop, colored by the amount of error in the resulting prediction. We see that for a subset of the initial conditions, two parameters can be found such that the predicted and measured impulses are equal, but such alignment does not hold globally, indicating that a significant number of impacts exist for which the measured impulse does not lie within the polygonal predictive range of the models. Fig.~\ref{fig:3.6}b shows the distribution of the model parameters found for each drop, and while the mean of these distributions is very close to the values in Tab. \ref{tab:3.1}, we see a significant amount of variance. The data suggests that: i) the analytical models are not expressive enough to explain all impacts (the predictive ranges do not cover a large enough region in the feasible impulse space), and ii) the choice of contact parameters should depend on the initial conditions of the impact, which is rarely done in practice.

\section{Data Driven Contact Models} \label{sec:data-driven-models}
Sec.~\ref{sec:3.2} highlights the variability in optimal contact parameters across experiments, and indicates that a significant portion of the imparted impulses do not lie in the predictive range of the commonly used contact models. This is in part due to the fact that the analytical contact models rely on parametrization of the impulse space and introduce additional constraints to limit the set of admissible imparted impulses. As \citet{chatterjee1997rigid} discusses, the parameters used are only very coarse approximations to very complicated and difficult to model phenomena, and there is no reason to expect that these or the constraints imposed on the interactions are physically meaningful.

In light of this, we propose two data-driven contact model formulations. The first class is purely data-driven and uses Gaussian Processes (GPs) trained on a feature space inspired by the analytical rigid contact models. The second class uses data to reinforce analytical models, either by informing parameter choice or correcting the predictions made by the models. The first class has weak priors, whereas the second class uses the analytical models which provide a much stronger prior.

\subsection{Class I - Data-Driven Contact Models} \label{sec:data-driven}
A contact model representation (Eq.~\ref{eq:2.1}) can be learned directly from the data using the feature space:
\begin{align}
    \mathcal{X} = \{m,I,r_x,r_y,\vect{q},\vect{v}\}^i \; \rightarrow \; \mathcal{Y} = \{ \vect{q}, \vect{v} \}^f
\end{align}
Using this feature space gives the learned models a large amount of flexibility at the cost of dimensionality of the space (10-D). For impacts that are near rigid we can exploit the structure and assumptions of rigid body contact to reduce the dimension of the feature space. The Energy Ellipse demarcates the region in space for all feasible imparted impulses under the rigid-body assumption that satisfy the law of conservation of energy. This space is a function of the effective inertia at the contact point ($\mat{M}_c$, square symmetric $2\times 2$ matrix) which captures the position of the center of mass with respect to the contact point as well as the inertia of the object, and the velocity of the contact point $\vect{v}_c$. As such we propose two feature spaces that exploit this structure:
\begin{align}
    \mathcal{X}_1 &= \{\mat{M}_{c,11},\mat{M}_{c,12},\mat{M}_{c,22},\vect{v}_{c,t},\vect{v}_{c,n}\} \\
    \mathcal{X}_2 &= \{(\mat{M}_c \vect{v}_c)_t,(\mat{M}_c \vect{v}_c)_n\} \\
    \mathcal{Y}_1 &= \{P_t, P_n \}
\end{align}
Here $\mathcal{X}_1$ and $\mathcal{X}_2$ are respectively 5- and 2-D, a significant reduction from the 10-D space. $\mathcal{X}_1$ attempts to capture the key features required to reconstruct the space of admissible impulses, while $\mathcal{X}_2$ is a coarse approximation of this space, where we consider the aggregate effect of inertia and velocity as the linear momentum of the contact point. We refer to these two models as ``\textbf{Data-driven Rigid Contact Models}'' due to their similarity to the analytical models.

The data-driven rigid contact models inherit the same properties as the analytical models, they assume a point contact that occurs instantaneously which results in an instantaneous linear impulse. Real objects often undergo small deformations during contact over a finite time horizon were forces are distributed over a finite area. We can allow greater expressibility of our contact models by permitting an instantaneous wrench (force and torque) prediction, i.e. using $(\mathcal{X}_1,\mathcal{Y}_2 = \{P_t, P_n, \tau \})$. We refer to this model as ``\textbf{Data-driven Contact Model}'' since this model deviates from the rigidity assumption. 

To learn these contact models we used Gaussian Processes (GPs) (\cite{rasmussen2006gaussian}) with an Automatic Relevance Determination (ARD) Square Exponential Kernel. We use this kernel to compensate for the relative differences in magnitudes of the features.

\subsection{Class II - Data Reinforced Contact Models} \label{sec:data-reinforced}
The models presented in sec.~\ref{sec:data-driven} are able to make predictions that span a larger subspace of the Energy Ellipse than their analytical counterparts, but can conceivably produce infeasible predictions until trained on a sufficient number of data.

In contrast analytical models, by construction, produce feasible predictions for reasonable choices of parameters. We can use data to reinforce these models and to this end we propose two formulations: i) using an analytical model and learning the residual errors in prediction using a GP, and ii) learning the optimal choice of model parameters for an analytical model given pre-impact states. Analogous to the naming convection in sec.~\ref{sec:data-driven} we will refer to these models as ``\textbf{Data-reinforced Rigid Contact Models}''. The former type of model is amenable to using an instantaneous wrench (i.e. $\mathcal{Y}_2$) and we will refer to the resulting model as ``\textbf{Data-reinforced Contact Model}''. In both cases we use the feature space $\mathcal{X}_5$ and the same kernels as sec.~\ref{sec:data-driven}.

\subsection{Data-driven Model Performance}

\begin{figure}
    \centering
    \includegraphics[width=0.99\textwidth]{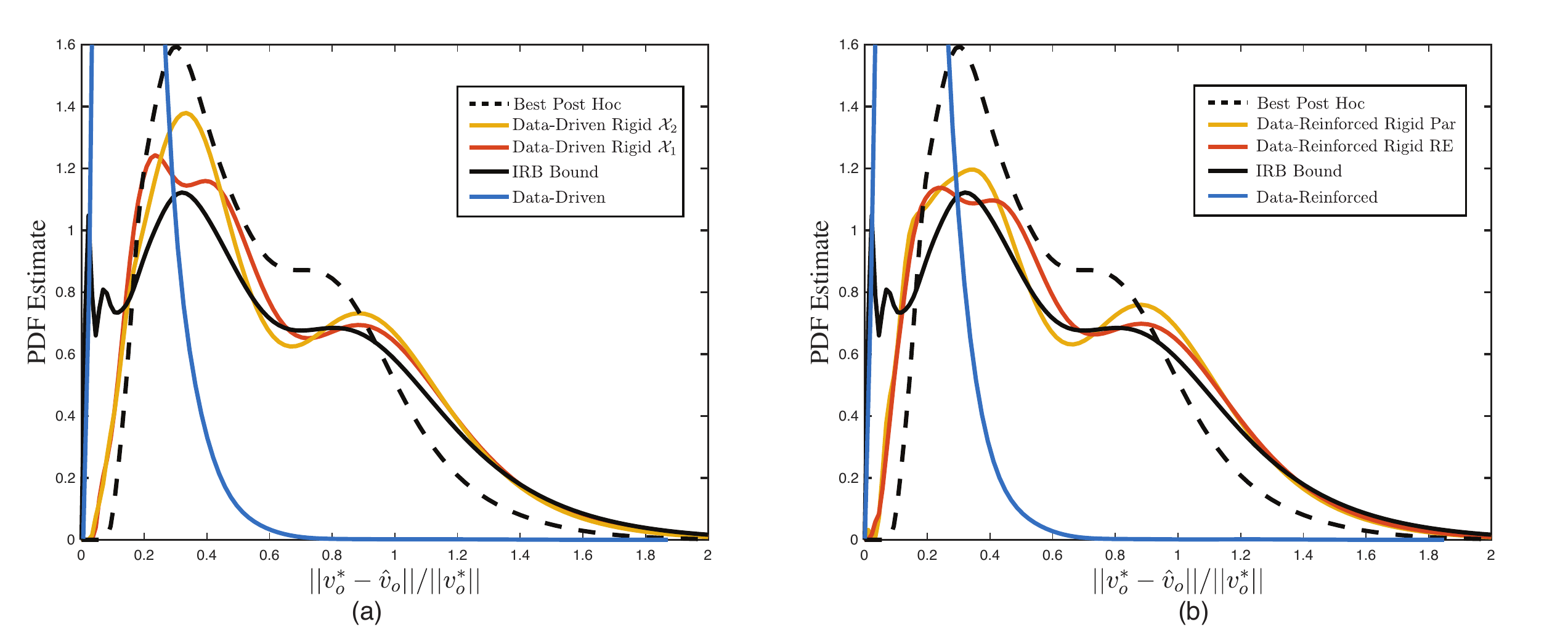}
    \caption{The estimated probability density function of the $\ell_2$ norm error in predicted post contact velocity for 450 data samples used to train the models. a) Class I, b) Class II.}
    \label{fig:4.1}
\end{figure}

Fig.~\ref{fig:4.1} shows the $\ell_2$ norm error in predicted post contact velocity of the data-driven models compared to the Best Post Hoc model (best performance of the analytical models) and the IRB Bound model (best performance of any rigid-body contact model). The rigid models of either class are able to outperform the Best Post Hoc model, but only approach the IRB Bound model. These models are able to outperform their analytical counterparts, but because they are built using the same features and assumptions as the analytical models, they will not be able to exceed the IRB bound. In Class I, the $\mathcal{X}_1$ shows a marginally better performance over $\mathcal{X}_2$, as it has a richer description of the impact event. The learned models with the extra explanatory power of wrench are able to outperform the IRB bound significantly and make far better predictions as they are not restricted by the same assumptions. 

\begin{figure}
        \centering
        \includegraphics[width=0.6\textwidth]{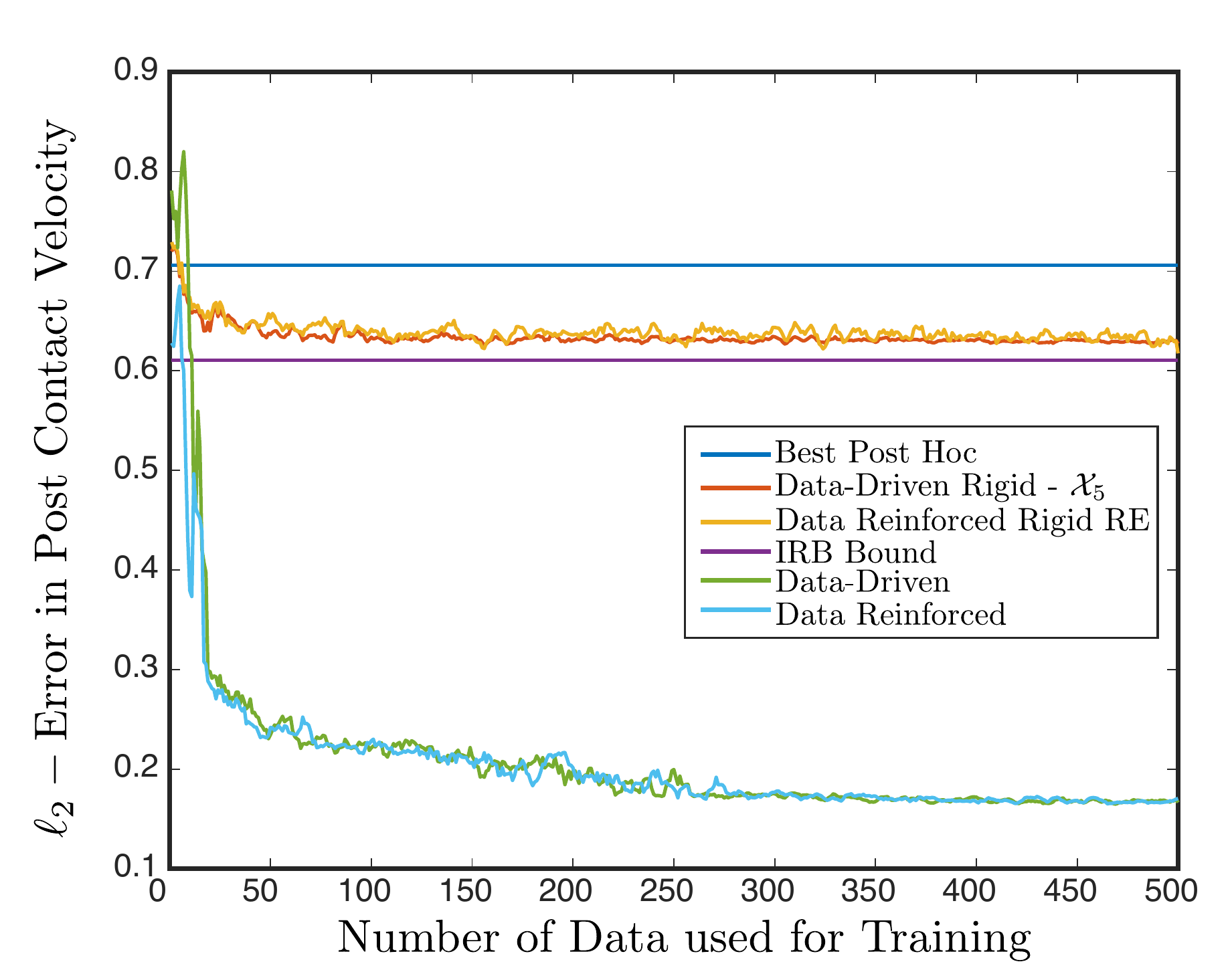}
        \caption[]{The $\ell_2$ norm error in predicted post contact velocity for the models studied vs. the number of data samples used to train models.}
        \label{fig:4.2}
\end{figure}

Fig.~\ref{fig:4.2} shows the expectation of the $\ell_2$-norm error in predicting the center of mass velocity as a function of the number of data used in the training, and here we show the best performing of each type of model for clarity. The data-driven and data-reinforced contact models are able to consistently and significantly outperform the IRB Bound due to their added expressibility, but require more data to converge (approximately 300 samples). The rigid learned models tend to converge to their optimal performance in approximately 100 samples, which is about double the number of samples required for the reliable identification of the parameters of the analytical models.

\section{Discussion \& Conclusion} \label{sec:conc}
Contact models fulfill an important role in robotics and are used extensively in locomotion \cite{kuindersma2016optimization} and manipulation \cite{ChavanDafle2015a}, yet the literature contains little work in empirical evaluation and comparison of the models prevalent in the field \cite{todorov2012mujoco}. In this paper we show that empirical data plays an important role in highlighting the limitations of analytical models, and in allowing the construction of data-driven contact models that outperform their analytical counter-parts. We also demonstrate that by exploiting known structure and assumptions about physical interactions, we can build models that are more data-efficient. In particular, here we were able to reduce a 10-D space to 5-D. We further show significant improvements in model prediction when we allow models do deviate slightly from the requirements of rigid contact, i.e. using instantaneous wrench instead of linear impulse. 

In this study we considered two classes of models, the purely data-driven (weak priors), and the data reinforced models (with strong priors). The challenge with using a purely data-driven model is the potentially poor or infeasible predictions made by these models before being trained on enough data. In contrast, augmenting analytical contact models with data allows for feasible predictions with adjustments made as more data is collected. With no data, the predictions rely on the prior generated by the underlying analytical models with no adjustment, which will be feasible but may be inaccurate. Our results suggest that these two formulations saturate at similar performances for this experimental setup, and the data reinforced models may be preferred due to the strong priors they provide.

The present study focuses on the impact of a single object, and learns a contact model suitable for the predictions of outcomes. Two interesting questions to ask are: how well does the model learned for this object work in making predictions for other objects of different shapes and physical properties? Is it possible to use this model as a prior and learn a correcting model with far fewer samples?

\clearpage


\bibliography{CoRL_2017_contact_modeling}
\end{document}